\documentclass{ifacconf}

\usepackage{graphicx}      
\usepackage{natbib}        
\usepackage{amsmath}       
\usepackage{comment}       
\usepackage{stfloats}
\usepackage{xcolor}

\begin{document}
\begin{frontmatter}

\title{Procedural Knowledge Extraction from Industrial Troubleshooting Guides Using Vision Language Models} 

\thanks[footnoteinfo]{Funding for the research was provided through Horizon Europe project AIXpert: (ID:101214389)}

\author{Guillermo Gil de Avalle,} 
\author{Laura Maruster,}
\author{Christos Emmanouilidis}

\address{University of Groningen, 9747 AE Groningen, The Netherlands \\ (g.gil.de.avalle; l.maruster; c.emmanouilidis)@rug.nl}

\begin{abstract}                
Industrial troubleshooting guides encode diagnostic procedures in flowchart-like diagrams where spatial layout and technical language jointly convey meaning. To integrate this knowledge into operator support systems, which assist shop-floor personnel in diagnosing and resolving equipment issues, the information must first be extracted and structured for machine interpretation. However, when performed manually, this extraction is labor-intensive and error-prone. Vision Language Models offer potential to automate this process by jointly interpreting visual and textual meaning, yet their performance on such guides remains underexplored. This paper evaluates two VLMs on extracting structured knowledge, comparing two prompting strategies: standard instruction-guided versus an augmented approach that cues troubleshooting layout patterns. Results reveal model-specific trade-offs between layout sensitivity and semantic robustness, informing practical deployment decisions.
\end{abstract}


\end{frontmatter}

\section{Introduction}
Industrial maintenance is a knowledge-intensive activity in which operators must diagnose faults, verify conditions, and select corrective actions under time pressure~\citep{roham2024knowledge}. Troubleshooting guides support this work by encoding domain expertise as step-by-step diagnostic procedures, commonly referred to as 'Procedural knowledge' (PK), through branching decision points, symbolic notations, and dense technical language~\citep{celino2025procedural}. Despite this valuable content, these documents are rarely adapted for machine consumption~\citep{Carriero2024,celino2025procedural}. Their heterogeneous layouts, inconsistent terminology, and reliance on visual relationships present significant challenges for automated processing~\citep{moritz2024review}.

Leveraging this expertise in operator support systems, i.e. digital tools that assist shop floor personnel during troubleshooting, requires extracting and structuring the procedural content. Such systems enable capabilities, including search over maintenance documentation, automated fault diagnosis, and integration of procedures with sensor data~\citep{roham2024knowledge}. While manual extraction is labor-intensive and error-prone, automated approaches are hindered by the documents' reliance on visual conventions and spatial relationships, creating a bottleneck in deploying these systems at scale.

Recent advances in Vision Language Models (VLMs) present a potential solution. Building on Large Language Models (LLMs), that is, neural models trained on massive text corpora \citep{Carriero2024}, VLMs also enable recognition and interpretation of visual information, making them well-suited for diagram-heavy documents. However, their effectiveness on industrial PK remains underexplored. Prior work on PK extraction has focused primarily on text-based instructional content such as household tasks, recipes, or textual maintenance instructions, where procedural structure is conveyed through linguistic patterns and sequential discourse~\citep{zhou2019learning, zhang2022reasoning, Carriero2024}. This leaves a gap in understanding whether VLMs can handle the spatial complexity and domain-specific conventions of industrial troubleshooting diagrams.

This paper aims to narrow this gap by evaluating two open-weight VLMs, namely, Pixtral-12b \citep{mistral2024pixtral} and Qwen2-VL-7b \citep{alibaba2024qwen2vl}, on the task of extracting procedural structured knowledge from industrial troubleshooting diagrams. Our approach introduces a small schema for uniform evaluation, capturing conditions, actions, and transitions to enable uniform evaluation across models. We compare two prompting strategies: a standard instruction-guided schema-based prompt and an augmented prompt that describes diagram visual conventions. By quantifying how model architecture and prompting affect extraction quality, this work provides actionable insights for deploying VLM-based knowledge extraction, with the ultimate goal of enabling operator support systems that can access and reason over industrial troubleshooting guides.

\section{Related Work}
\subsection{Procedural Knowledge in Industry}
Current methods for automatic PK extraction focus largely on text-rich domains, where linguistic patterns convey procedural structure. Household task instructions demonstrate this paradigm: procedural steps can be learned through embeddings that capture step relevance and temporal ordering from WikiHow articles~\citep{zhou2019learning}. Broader surveys confirm that extraction methods across procedural domains rely on linguistic markers such as imperative verbs, temporal connectives, and sequential discourse patterns~\citep{zhang2022reasoning}. Recent applications of LLMs extend this approach to maintenance manuals and recipes, using prompt engineering to extract procedural steps, actions, and equipment~\citep{Carriero2024}, yet these methods process only stepwise textual content extracted from PDFs, explicitly removing visual layout and formatting.

Industrial troubleshooting guides operate under different principles. Diagnostic logic is often encoded spatially through flowchart conventions: decision points appear as diamond shapes, actions as rectangles, and flow direction as arrows~\citep{moritz2024review}. Textual labels encoding conditions, outcomes, or observations are often minimal, relying on visual structure to convey procedural logic. While semantic markers such as numbering or reference codes may appear, their spatial distribution across the page can disrupt sequential interpretation when elements are processed independently.

The cognitive demands of interpreting such documents are substantial even for human experts. Studies of maintenance technicians working with schematic and wiring diagrams show that tracing components across pages causes significant loss of contextual information, requiring specialized visualization support including highlighting, spatial context preservation, and distortion techniques to maintain orientation in prolonged use  ~\citep{woo2009sdviz,kim2010evaluating}. These findings suggest that the visual complexity of technical diagrams is not merely an inconvenience but a fundamental barrier to efficient interpretation, whether by humans or automated systems.

Moreover, manual extraction of PK from such documents is labor-intensive and error-prone: domain experts must interpret each diagram and encode its contents into structured formats, a process that does not scale as equipment portfolios expand and documentation accumulates~\citep{celino2025procedural}. Rule-based automation offers limited relief, as individual templates must be curated per manufacturer and document type, precluding generalization across heterogeneous layouts~\citep{moritz2024review}. This creates a bottleneck: procedural knowledge that could enable operator support systems remains locked in visual formats that resist systematic extraction.

\subsection{Vision Language Models for Documents}

Vision Language Models offer a potential solution by jointly processing visual layout and textual content. Models such as GPT-4V~\citep{openai2023gpt4} and open-weight alternatives like LLaVA~\citep{liu2023visual} have demonstrated strong performance on tasks requiring multimodal reasoning, including visual question answering and image captioning. For document understanding specifically, VLMs can interpret typography, spatial arrangement, and graphical elements alongside text, capabilities that align well with the requirements of troubleshooting diagram extraction.

Recent work has begun exploring VLMs for diagram comprehension. ~\citet{pan2024flowlearn} introduce FlowLearn, a benchmark for evaluating VLM performance on flowchart understanding, demonstrating that while current models can recognize visual components and perform text extraction, they struggle with complex structural reasoning tasks such as node counting and flow logic interpretation. Deep learning approaches for engineering diagram digitization~\citep{moritz2024review} have shown that convolutional and graph neural networks can extract structural information by learning associations between visual shapes and semantic roles. Chart question answering systems~\citep{masry2022chartqa} demonstrate that VLMs can decode visual encodings of quantitative data when prompted appropriately. These results suggest that VLMs possess the architectural capacity to handle spatial reasoning tasks, but evaluation has concentrated on general-domain benchmarks like DocVQA~\citep{mathew2021docvqa} and chart datasets where symbology is standardized and publicly documented.

Industrial troubleshooting diagrams introduce challenges not present in these benchmarks. Terminology is domain-specific, often referencing equipment codes, fault modes, and operational parameters unfamiliar to general-purpose models~\citep{Carriero2024,roham2024knowledge}. Visual conventions vary across manufacturers and evolve with equipment generations, meaning that shape-to-function mappings are not universally consistent. Diagram quality varies as well: older documents may be scanned from physical copies with degraded legibility, while newer guides may embed complex nested structures that require complex reasoning to interpret correctly~\citep{moritz2024review}. Whether VLMs can handle these industrial-specific characteristics for PK extraction remains an open question.

The models evaluated in this paper, Pixtral-12B~\citep{mistral2024pixtral} and Qwen2-VL-7B~\citep{alibaba2024qwen2vl}, represent recent advances in open-weight VLM architectures with different design choices. Pixtral employs a vision transformer that uses cross-attention mechanisms to link visual elements with text tokens. This allows the model to jointly process images and text by computing attention weights that determine which visual regions are relevant to which text elements. Qwen2-VL uses a dynamic resolution strategy that automatically adjusts how the image is processed based on content density. When encountering text-heavy regions, the model increases processing granularity to preserve fine-grained details that might otherwise be lost at lower resolutions. To the authors' knowledge, neither model has been systematically evaluated on industrial PK extraction. This work addresses that gap by introducing a specific evaluation grounded in the requirements of the industrial domain, comparing how model architecture and prompting strategy affect extraction quality on proprietary troubleshooting diagrams.

\section{Methodology}

\subsection{Dataset}
\begin{figure}[ht]
    \centering
    \includegraphics[width=0.8\columnwidth]{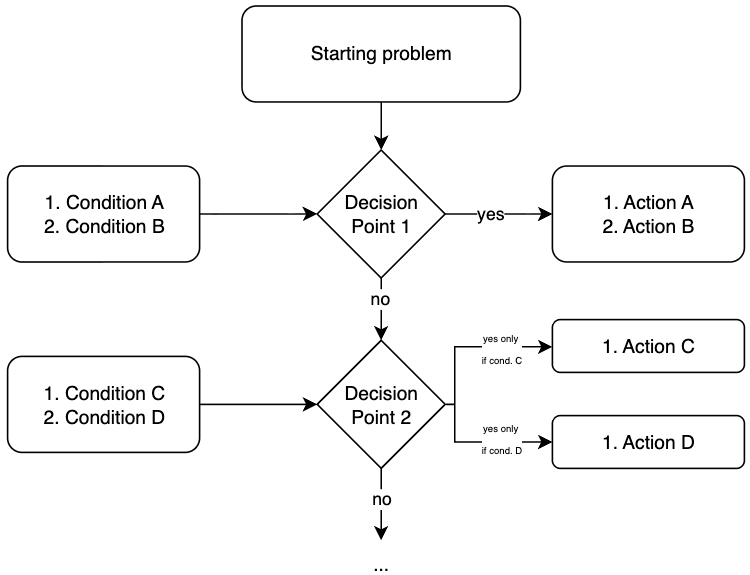}
    \caption{Example of troubleshooting guide structure.}
    \label{fig:guide}
\end{figure}

The study uses proprietary industrial troubleshooting guides written in Dutch, provided by a local manufacturer. Each of the 12 provided guides consists of one to three pages (24 pages total) and follows a flowchart-like style. Rectangular boxes describe observations or actions, diamond shapes represent decision points, and arrows define the procedural flow, as exemplified in Figure~\ref{fig:guide}. Arrows connected to decision nodes include the labels “ja” or “nee” (“yes” or “no”) to indicate the correct branch. Each document contains roughly 30 to 100 entities (i.e., individual conditions, decisions, or actions) and 30 to 60 relationships.

\begin{figure*}[b]
    \centering
    \includegraphics[width=\textwidth]{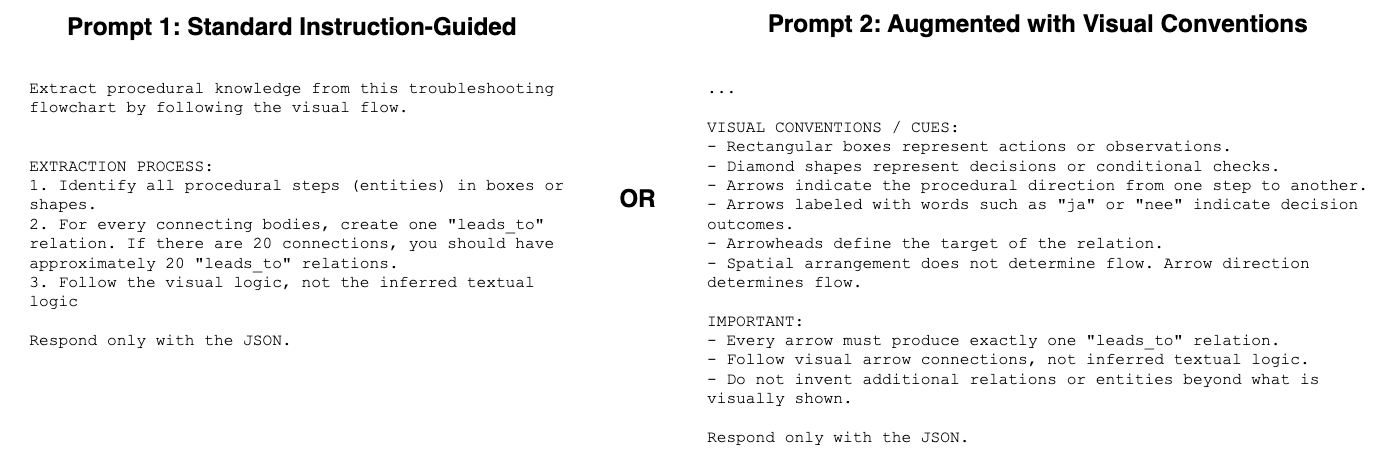}
    \caption{Prompting strategies}
    \label{fig:prompt}
\end{figure*}

\subsection{Extraction Task}
The extraction task requires each model to identify procedural entities and reconstruct the relations that link them into a coherent process structure. Our extraction schema uses vocabulary aligned with the visual structure of the provided guides themselves, as seen in Figure~\ref{fig:guide}. These diagrams follow a consistent pattern where observations lead to decision points, which branch into one or more sets of prescribed actions. This structure is also consistent with established procedural ontologies, such as P-PLAN~\citep{garijo2012pplan}.

As shown in Table~\ref{tb:schema}, the schema defines three entity types (\texttt{Condition, Action, Decision}) and one relation type (\texttt{isPreceededBy}). Conditions represent observations or states to be verified, Decisions indicate branching points with yes/no outcomes, and Actions specify operations to be performed. This schema provides a uniform representation across all guides and enables consistent comparison across VLMs. The task is framed as a structured output generation problem, in which a model receives a page-level image and returns a list of entities and relations expressed in a predefined JSON format.

\begin{table}[ht]
\begin{center}
\caption{Schema for entities and relations}\label{tb:schema}
\begin{tabular}{p{2.7cm} p{4.2cm}}
\hline
\textbf{Entity Type} & \textbf{Description} \\
\hline
Condition & A condition or state to be verified \\
Action & An operation to be performed \\
Decision & A decision point with branching outcomes \\
\hline
\textbf{Relation Type} & \textbf{Description} \\
\hline
isPreceededBy & One step is preceded by another in the procedure \\
\hline
\end{tabular}
\end{center}
\end{table}

\subsection{Prompting Strategies}
Each model was tested with two prompting strategies, as seen in Figure~\ref{fig:prompt}. The first strategy uses a standard instruction-guided prompt, providing only initial instructions, the schema outline, and an example of the JSON output for reference. The second strategy extends the prompt with explicit descriptions of the visual conventions used in the diagrams, including the functional role of shapes and the interpretation of arrows and branch labels. 

These strategies were chosen to evaluate whether explicit domain knowledge improves extraction quality, following ~\citet{Carriero2024}, which demonstrated that including concise entity definitions significantly improved LLM extraction accuracy compared to both verbose definitions and no definitions. Similarly, we hypothesize that providing explicit visual convention descriptions may help VLMs correctly interpret flowchart symbols, which unlike text-based procedures require understanding spatial relationships and symbolic notation. The standard prompt serves as a baseline to isolate the effect of visual domain-specific guidance.

\subsection{Extraction Procedure}
All models were executed locally on a high-performance computing cluster equipped with one Nvidia A100 GPU accelerator card with 40 GB of VRAM. Each page within the guides was processed independently. For each model and prompting strategy, the system returned a structured inference containing entity candidates and their relations. Outputs were parsed automatically, and models that failed to return valid JSON were treated as producing zero entities and relations. No fine-tuning or post-training adaptation was applied. Text normalization was applied to both predicted and ground truth entities to enable fair comparison: we removed numerical prefixes (e.g., ``4-0'', ``A1)''), punctuation, and excess whitespace to focus evaluation on semantic content rather than formatting artifacts. No semantic correction was introduced, so as to reflect the inherent extraction capabilities of each model.

\subsection{Evaluation}
All twelve guides were manually annotated by domain experts following the structure of Table~\ref{tb:schema}, generating a domain-specific gold standard comprising 548 entities and 536 relations.  For illustration, an annotated excerpt might look as follows, with the original text written in Dutch:

\begin{verbatim}
E1: "Check water pressure" (type: Condition)
E2: "Open valve" (type: Action)
R1 : E1 isPreceededBy E2
\end{verbatim}

Entity matching employed text comparison with a similarity threshold of 0.9, computed over exact token overlap between entity texts after lemmatization. Entity types were required to match exactly. Relation matching followed a content-based strategy: an inferred relation was considered correct if both its source and target entities matched ground truth entities and the relation type was identical, regardless of entity identifiers (which vary between model outputs but do not reflect semantic differences).

Performance was evaluated using standard metrics computed separately for entities and relations. \textit{Entity precision} measures the proportion of extracted entities that are correct, \textit{entity recall} measures the proportion of ground truth entities successfully extracted, and \textit{entity F1} is their harmonic mean. \textit{Relation precision}, \textit{relation recall}, and \textit{relation F1} are defined analogously for extracted relations. True positives, false positives, and false negatives were aggregated across all documents to compute overall metrics.

\section{Results}

\begin{table*}[t]
\centering
\caption{Aggregate Performance}\label{tb:aggregate}
\begin{tabular}{llcccccc}
\hline
\textbf{Model} & \textbf{Prompt} & \textbf{Ent. Precision} & \textbf{Ent. Recall} & \textbf{Ent. F1} & \textbf{Rel. Precision} & \textbf{Rel. Recall} & \textbf{Rel. F1} \\
\hline
Qwen2-VL-7B & Standard & 0.305 & 0.383 & \textbf{0.340} & 0.077 & 0.050 & 0.061 \\
Qwen2-VL-7B & Augmented & 0.203 & 0.414 & 0.272 & 0.177 & 0.076 & \textbf{0.107} \\
Pixtral-12B & Standard & 0.336 & 0.263 & 0.295 & 0.018 & 0.013 & 0.015 \\
Pixtral-12B & Augmented & 0.383 & 0.173 & 0.239 & 0.008 & 0.004 & 0.005 \\
\hline
\end{tabular}
\end{table*}

We evaluated Qwen2-VL-7B and Pixtral-12B on procedural knowledge extraction from 12 industrial troubleshooting guides (24 pages total) under standard and augmented prompting conditions. Table~\ref{tb:aggregate} presents aggregate performance. Both models demonstrated limited extraction capabilities, with entity F1 scores ranging from 0.24 to 0.34 and relation F1 scores below 0.11. Qwen2-VL achieved the highest entity extraction performance (F1 = 0.340 under standard prompting), while both architectures exhibited severe deficits in relation extraction, the critical component for PK to be considered as such.

The models exhibited distinctly different performance patterns. Qwen2-VL demonstrated high variance across documents, with per-guide entity F1 scores ranging from 0.00 to 0.78 under standard prompting, revealing a threshold effect where documents either achieved moderate success or failed catastrophically. Pixtral showed more consistent but universally limited performance, with most documents in the 0.24 to 0.44 F1 range. Table~\ref{tb:variance} illustrates this variance through selected examples spanning high-performing, moderate, and failed extractions.

Prompt engineering produced divergent effects. For Qwen2-VL, augmented prompting relatively improved relation extraction (F1: 0.061 to 0.107) while reducing entity precision (0.305 to 0.203), suggesting that explicit structural guidance redirects attention toward graph connectivity at the expense of entity accuracy. Conversely, Pixtral-12B showed degradation under augmented prompting for both entities (F1: 0.295 to 0.239) and relations (F1: 0.015 to 0.005), indicating that increased prompt complexity may conflict with its processing architecture.

\begin{table}[ht]
\begin{center}
\caption{Per-Document Performance Variance}\label{tb:variance}
\begin{tabular}{llcccc}
\hline
\textbf{Guide} & \textbf{Model} & \textbf{Prompt} & \textbf{Ent F1} & \textbf{Rel F1} & \textbf{TP/FP} \\
\hline
27 & Qwen & Std & 0.657 & 0.138 & 22/6 \\
19 & Pixtral & Aug & 0.778 & 0.000 & 14/1 \\
17 & Pixtral & Std & 0.443 & 0.035 & 27/35 \\
4 & Qwen & Std & 0.152 & 0.000 & 15/132 \\
26 & Pixtral & Aug & 0.000 & 0.000 & 0/0 \\
\hline
\end{tabular}
\end{center}
\end{table}

Relation extraction emerged as the primary bottleneck. Under standard prompting, Qwen2-VL extracted only 27 true positive relations from 536 ground truth (5\% recall) while generating 324 false positives. Pixtral-12B performed worse, with 7 true positives against 381 false positives. Critically, 9 of 12 documents yielded zero correctly extracted relations for Qwen under standard prompting, and 11 of 12 for Pixtral under augmented prompting, rendering outputs unsuitable for procedural graph reconstruction.

\section{Discussion}

The observed performance limitations stem from distinct failure modes that manifest differently across architectures, revealing fundamental challenges for VLM deployment in industrial document understanding.

\textbf{Failure mode analysis} Qwen2-VL's results reflect a failure mode not commonly reported in VLM literature~\citep{mathew2021docvqa,masry2022chartqa}, which we refer to as \textit{infinite loop collapse}. Documents that failed completely (F1 $\leq$ 0.15) exhibited systematic hallucination: after extracting 15-25 correct entities, the model entered repetitive loops generating hundreds of near-identical entities with incrementing identifiers. One guide produced 132 false positives versus 15 true positives, with sequences like ``E16: Zijn er verkeerde doppen aanwezig?'', repeated every time with different identifier. The triggering mechanism appears sensitive to factors beyond document structure, as all guides follow identical flowchart-like layouts. Documents avoiding the infinite loop maintained low false positive rates (5-6 FP) while achieving 51-79\% entity recall, demonstrating adequate capacity when this failure mode is suppressed.

Pixtral exhibited different pathologies. Despite its larger size (12B vs 7B parameters of Qwen2-VL), Pixtral achieved only 17-26\% entity recall compared to Qwen's 38\% under comparable conditions and extracted almost no valid relations. The model generated fewer false positives (mean 24 FP vs Qwen's 40), suggesting better calibrated stopping criteria, yet this came at the cost of missing most entities. The performance gap appears to be rooted in how each model processes visual information: Pixtral focuses attention between visual elements and text tokens~\citep{mistral2024pixtral}, while Qwen dynamically adjusts image resolution to preserve fine-grained detail~\citep{alibaba2024qwen2vl}. Qwen's approach enables more complete parsing of dense flowcharts, whereas Pixtral appears to struggle when tracking the overlapping arrows and tightly packed blocks of Dutch text that characterize these diagrams. This aligns with findings from FlowLearn~\citep{pan2024flowlearn}, where VLMs succeeded in recognizing individual components but failed in interpreting flow logic.

\textbf{Relation extraction bottleneck.} Both models achieved relation F1 below 0.11, with most documents producing zero correctly extracted relations despite containing 35 to 61 ground truth relations. This failure extends beyond entity errors: documents with moderate entity F1 (0.40 to 0.66) still yielded zero relation F1, indicating that models often identified nodes but failed to parse the connecting arrows. Two mechanisms may contribute. First, token budget misallocation, where the model expends most of its output on enumerating entities rather than allocating space for relations, leads to truncated extractions. In several cases, long sequences of hallucinated entities consumed the entire output window. Second, parsing spatial relationships in diagrams with overlapping arrows and densely packed nodes exceeds current architectural or training capabilities. Augmented prompting partially addressed the first mechanism for Qwen2-VL by requesting graph structure early, but the same approach failed for Pixtral and still left Qwen2-VL far below industrial requirements, suggesting that limitations in spatial reasoning remain dominant~\citep{moritz2024review}.

\textbf{Architectural implications.} The models' response to different prompt strategies reveals fundamental capability differences. Qwen's improved relation extraction under augmented prompting suggests latent capacity to identify graph structure when explicitly directed, but insufficient instruction-following robustness to prioritize this task without guidance. The precision-recall tradeoff (entity precision declined while relation precision improved) indicates that the model reallocates its limited attention and computation toward whichever aspect the prompt highlights, rather than balancing both tasks based on actual requirements. This reflects the limited instruction-following robustness, typical of smaller models. Conversely, Pixtral's degradation under augmented prompting (both entity and relation F1 declined) points to capacity saturation where detailed instructions overload context processing. The absence of any improved dimension suggests architectural rather than prompt-correctable limitations.

\textbf{Addressing hallucination in industrial contexts.} The infinite loop collapse Qwen2-VL could be addressed by targeted mitigation strategies. Constrained decoding offers potential: implementing runtime validators that detect repetition patterns (e.g., identical text with incrementing numbers appearing more than three times) and terminate generation could prevent token budget exhaustion while preserving successful extractions. This approach requires no model retraining. Iterative extraction could also provide a complementary strategy by dividing diagrams into spatial regions (e.g., quadrants or sequential process stages) and processing each independently before merging results with duplicate detection. This could reduce the working memory demands that appear to trigger infinite loop collapses when models attempt to process all entities simultaneously.

Domain-specific fine-tuning represents a medium-term solution for organizations with access to larger internal documentation. The consistent visual structure of troubleshooting guides makes them suitable for transfer learning with relatively modest training sets. Fine-tuning Qwen2-VL on a substantial number of annotated diagrams could improve spatial parsing, reduce hallucination by reinforcing completion recognition, and address instruction-following deficits that prompting alone cannot resolve. However, our twelve-guide dataset is insufficient for such training, reflecting the common challenge of limited proprietary data in industrial settings \citep{roham2024knowledge}and underscoring the need for collaborative data agreements or data augmentation approaches.

\textbf{Deployment considerations.} Current open-source VLMs remain unsuitable for autonomous deployment in safety-critical settings where extraction errors could propagate into maintenance decision systems~\citep{roham2024knowledge}. The combination of high entity miss rates, relation miss rates exceeding 90 percent, mismanagement of token budgets, missing stopping points, and unpredictable failure modes (different for Qwen and Pixtral) prevents reliable automation. Nevertheless, documents without hallucinations reached entity F1 scores between 0.48 and 0.78, indicating that models such as Qwen2-VL, whose major failure modes may be isolatable, could support human-in-the-loop workflows that pre-populate extraction interfaces and reduce annotation time, as suggested in prior work~\citep{Carriero2024}. Achieving this requires reliable hallucination detection to identify failed extractions, even before human review.

An open question remains on whether text-only LLMs, which do not process diagram layout directly but avoid visual failure modes, might achieve comparable benefits in human-in-the-loop workflows, given that relation extraction performance is already low in our experiment. Alternatively, if relation extraction in VLMs can be substantially improved, their ability to leverage spatial conventions could again make them the more attractive option. 

\section{Conclusion}

This work evaluates open-source VLMs for procedural knowledge extraction from industrial troubleshooting diagrams, benchmarking two architectures across multiple prompting strategies. Results demonstrate significant limitations, with entity F1 of 0.24-0.34 and relation F1 below 0.11, far short of requirements for autonomous deployment in industrial knowledge graph construction. Relation extraction represents a fundamental bottleneck not addressable through prompt engineering alone.

Model comparison revealed distinct architectural tradeoffs: Qwen2-VL-7B achieved higher peak performance (F1 up to 0.78) but exhibited infinite loop collapse in 40\% of documents, entering repetitive loops preventing relation extraction. Pixtral-12B demonstrated consistent behavior but suffered fundamental capacity limitations, extracting only 2-7 valid relations total despite larger parameter count. Augmented prompting improved Qwen's relation extraction 75\% while degrading Pixtral's performance, revealing model-specific optimization necessities.

Several limitations were identified in this study, including hardware constraints (40GB VRAM) preventing evaluation of larger architectures, the proprietary dataset precluding closed-source model comparison, and single-manufacturer Dutch-language documentation limiting cross-domain generalization. Moreover, the limited sample size (12 guides) precludes formal statistical significance testing, though the observed performance differences are substantial. Future work should explore larger model scales, domain-specific fine-tuning, and constrained decoding strategies to address the hallucination and relation extraction bottlenecks identified here.

This work quantifies the industrial AI readiness gap for maintenance documentation processing and provides validated datasets and evaluation methodologies. Near-term deployment should focus on human-AI collaboration where VLMs augment rather than replace expertise, with automation expanding as reliability improves through domain-specific fine-tuning and architectural innovation in spatial reasoning.

\begin{ack}
This research has been sponsored by the Horizon Europe project AIXpert (ID: 101214389). The authors wish to extend special thanks to Philips Lifestyle Consumer B.V., in particular Eric Sloot, for the valuable collaboration and support in this work.
\end{ack}

\section*{DECLARATION OF GENERATIVE AI AND AI-ASSISTED TECHNOLOGIES IN THE WRITING PROCESS}
During the preparation of this work the author(s) used Claude Sonnet 4.5 for coding and text refinement purposes. After using this tool/service, the author(s) reviewed and edited the content as needed and take(s) full responsibility for the content of the publication.

\bibliography{ifacconf}     
                                                   
\end{document}